\documentclass[11pt,a4paper]{article}
\usepackage[hyperref]{acl2019}
\usepackage{times}
\usepackage{latexsym}
\usepackage{graphicx}
\usepackage{amsfonts}
\usepackage{amsmath}

\usepackage{wasysym}
\usepackage{hyperref}
\usepackage{verbatim}

\usepackage{booktabs}
\usepackage{multirow}
\usepackage{url}

\usepackage[utf8]{inputenc}
\usepackage{array}
\usepackage{makecell}

\aclfinalcopy 

\title{Deep Contextualized Biomedical Abbreviation Expansion}
\author{Qiao Jin \\
  University of Pittsburgh \\
  {\tt qiao.jin@pitt.edu} \\\And
  Jinling Liu \\
  University of Pittsburgh \\
  {\tt jil172@pitt.edu} \\\AND
  Xinghua Lu \\
  University of Pittsburgh \\
  {\tt xinghua@pitt.edu} \\
  }

\date{}

\begin{document}

\maketitle
\begin{abstract}
Automatic identification and expansion of ambiguous abbreviations are essential for biomedical natural language processing applications, such as information retrieval and question answering systems. 
In this paper, we present DEep Contextualized Biomedical Abbreviation Expansion (\textbf{DECBAE}) model. 
DECBAE automatically collects substantial and relatively clean annotated contexts for 950 ambiguous abbreviations from PubMed abstracts using a simple heuristic.
Then it utilizes BioELMo \cite{jin2019probing} to extract the contextualized features of words, and feed those features to abbreviation-specific bidirectional LSTMs, where the hidden states of the ambiguous abbreviations are used to assign the exact definitions. 
Our DECBAE model outperforms other baselines by large margins, achieving average accuracy of $0.961$ and macro-F1 of $0.917$ on the dataset. It also surpasses human performance for expanding a sample abbreviation, and remains robust in imbalanced, low-resources and clinical settings.
\end{abstract}

\section{Introduction}
Abbreviations are shortened forms of text-strings. They are prevalent in biomedical literature such as scientific articles, clinical notes and user queries in information retrieval systems. Abbreviations can be ambiguous (e.g.: ER can refer to estrogen receptor, endoplasmic reticulum, emergency room etc.), especially when they appear in short or professional texts where the definitions are not given. For instance, about 15\% of PubMed queries include abbreviations \cite{islamaj2009understanding}, and about 14.8\% of all tokens in a clinical note dataset are abbreviations \cite{xu2007study}. In both cases, the definitions of the abbreviations are rarely provided. Thus, automatic expansion of ambiguous abbreviations to their full forms is vital in biomedical natural language processing (NLP) systems.

In this paper, we focus on the cases where definitions of ambiguous abbreviations are not directly available in the contexts, so reasoning over the contexts is required for disambiguation. Under the conditions where definitions are provided in the contexts, one can easily extract them using rule-based methods. 

We present DEep Contextualized Biomedical Abbreviation Expansion (DECBAE) model. DECBAE uses a simple heuristic to automatically construct large supervised disambiguation datasets for 950 abbreviations from PubMed abstracts: In scientific writing, authors define abbreviations the first time they are used, and the same abbreviations in the following sentences have the same definitions as those of the first ones. We extract all the sentences containing the same abbreviations in each PubMed abstract, and use the definition given in the first sentence as the full form label of abbreviations in the following sentences. We group the definitions for each abbreviation and formulate abbreviation expansion as a classification task, where input is an ambiguous abbreviation with its context, and the output is one of its possible definitions.

Recent breakthroughs of language models (LM) pre-trained on large corpora like ELMo \cite{peters2018deep} and BERT \cite{devlin2018bert} clearly show that unsupervised LM pre-training can vastly improve performance of downstream models. To fully utilize the knowledge encoded in PubMed abstracts, DECBAE uses BioELMo \cite{jin2019probing}, a domain adapation verison of ELMo, to embed the words. After the embedding layer, DECBAE applies abbreviation-specific bidirectional LSTM (biLSTM) classifiers to do the abbreviation expansion, where the biLSTM parameters are trained separately for each abbreviation. We train DECBAE from the automatically collected dataset of 950 ambiguous abbreviations.

At inference time, DECBAE feeds the BioELMo embeddings of the whole sentence and uses the corresponding abbreviation-specific biLSTM classifiers to perform disambiguation of abbreviations in the sentence. We show that DECBAE outperforms other baselines by large margins and even performs better than single human expert. Although training instances of DECBAE are collected from PubMed, it covers 85\% of clinically related abbreviations mentioned in a previous work \cite{xu2012combining}. Moreover, DECBAE remains robust in low-resource and imbalanced settings.

\section{Related Work}
\paragraph{Contextualized word embeddings:} Recently, contextualized word representations pre-trained by large corpora like ELMo \cite{peters2018deep} and BERT \cite{devlin2018bert} significantly improve the performance of various NLP tasks. ELMo is a pre-trained biLSTM language model. ELMo word embeddings are calculated by a weighted sum of the hidden states of each biLSTM layer. The weights are task-specific learnable parameters while biLSTM layers are fixed. In-domain trained contextual embeddings further improve the performance on domain-specific tasks. In this paper, we use BioELMo, which is a biomedical version of ELMo trained on 10M PubMed abstracts \cite{jin2019probing}. BioELMo outperforms general ELMo by large margins on several biomedical NLP tasks.

We don't use BERT for contextualized embeddings due to its fine-tuning nature: users just need to download 1 BioELMo and $N$ abbreviation-specifc biLSTM weights to run DECBAE locally, which takes significantly less disk size than $N$ fine-tuned BERTs for each abbreviation. $N$ is the number of abbreviations.

\paragraph{Word sense disambiguation (WSD):} The goal of WSD is to determine the correct sense of words in different contexts. Abbreviation expansion is a specific case of WSD where the ambiguous words are abbreviations. In this paper, we use abbreviation expansion and abbreviation disambiguation interchangeably. Several human-annotated datasets are available for supervised WSD \cite{navigli2013semeval,camacho2016nasari,raganato2017word}. However, human annotations could be expensive, especially in domain specific settings. To address this problem, some automatic dataset collection methods have been proposed \cite{yu2007using, ciosici2019unsupervised}, where abbreviations are automatically labeled if they are defined previously in the same documents. We use a similar approach in this work.

\citet{peters2018deep} report that just matching the ELMo embedding of the target words with the nearest sense representations, calculated by averaging their ELMo embeddings, leads to comparable WSD performance with state-of-the-art models using hand crafted features \cite{iacobacci2016embeddings} or task-specific biLSTM trained with multiple tasks \cite{raganato2017neural}. Instead of searching the nearest contextualized embeddings neighbors of the abbreviation and definitions, we model abbreviation expansion as classification.

\paragraph{Biomedical abbreviation expansion:} Various methods have been introduced for automatically expanding biomedical abbreviations. 
\citet{yu2007using} train naive Bayes and SVM classifiers with bag-of-word features on an automatically collected dataset from PubMed. Some works disambiguate abbreviations to their senses in controlled vocabularies like Medical Subject Headings\footnote{\url{https://www.nlm.nih.gov/mesh}} (MeSH) and Unified Medical Language System\footnote{\url{https://www.nlm.nih.gov/research/umls/}} (UMLS). \citet{xu2015clinical} use pooled neighbor word embeddings of the abbreviations as features to train SVM classifiers for clinical abbreviaiton disambiguation. \citet{jimeno2011exploiting} introduced MSH WSD dataset to test the performance of supervised biomedical WSD systems and several supervised models have been proposed on it \cite{antunes2017supervised, yepes2017word}. Recently \citet{pesaranghader2019deepbiowsd} presented deepBioWSD which sets new state-of-the-art performance on it. DeepBioWSD uses a single biLSTM encoder for disambiguation of all abbreviations by calculating the pairwise similarity between context representations and sense representations. 

To the best of our knowledge, DECBAE is the first model that uses deep contextualized word embeddings for biomedical abbreviation expansion.

\section{Methods}
\begin{figure*}
    \centering
    \includegraphics[width=\linewidth]{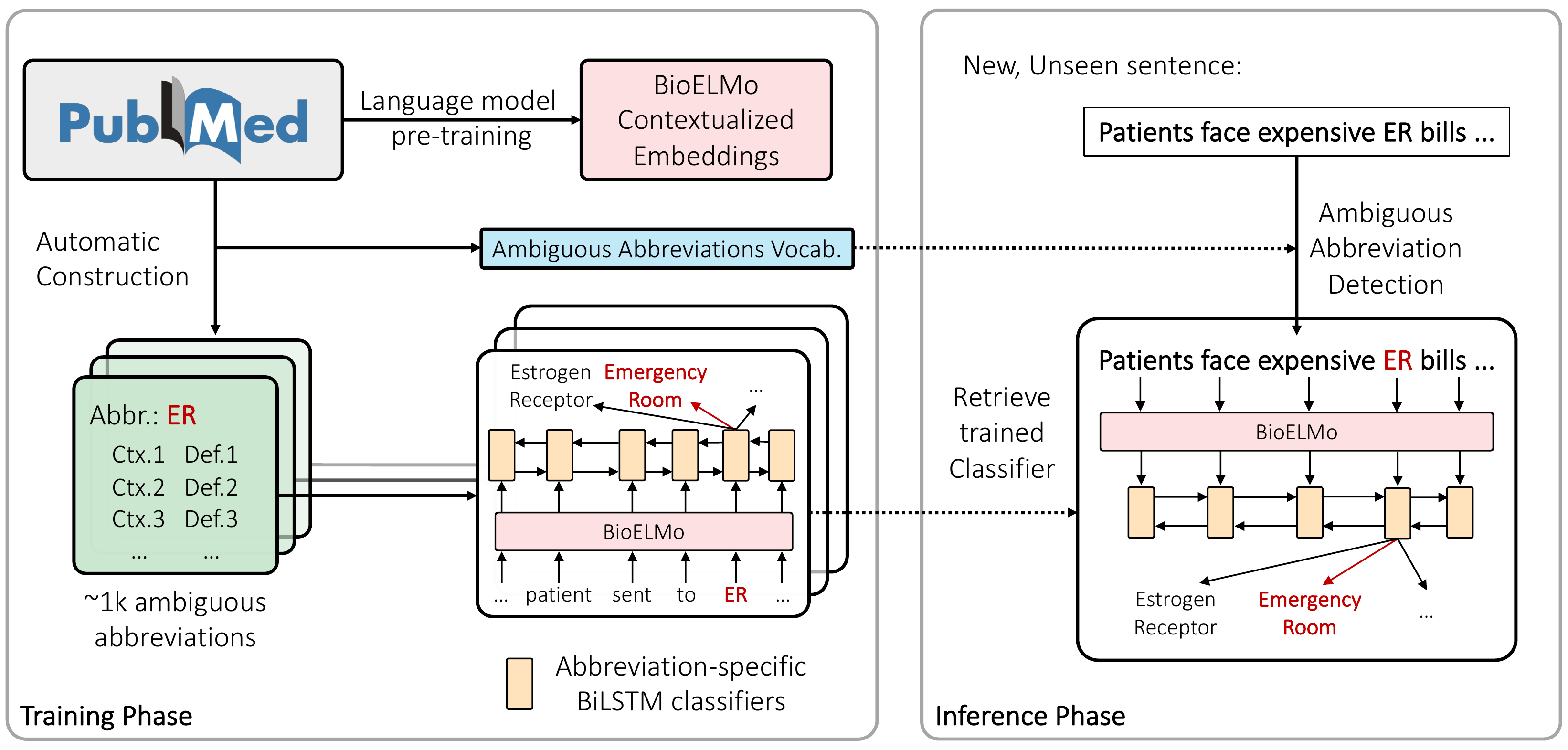}
    \caption{Architecture of DECBAE. Training and inference phases are illustrated in the left and right boxes, respectively. The PubMed corpus is used in training BioELMo \cite{jin2019probing} and collecting the disambiguation dataset. We train a separate biLSTM classifier for each abbreviation, and the specific pre-trained classifier is retrieved in inference phase.}
    \label{fig:pipeline}
\end{figure*}

Figure \ref{fig:pipeline} shows the architecture of DECBAE. During training, we first construct abbreviation expansion datasets from PubMed (\S\ref{constr}). We use BioELMo (\S\ref{bioelmo}) to get the contextualized representations of words, and train a specific biLSTM classifier (\S\ref{bilstm}) for each abbreviation. During inference (\S\ref{infer}), we first detect whether there are ambiguous abbreviations in input sentences by the expert-curated ambiguous abbreviation vocabulary. If so, we use BioELMo and the corresponding abbreviation-specific biLSTM classifiers to do the disambiguation.

\subsection{Dataset Collection} \label{constr}

\begin{figure*}
    \centering
    \includegraphics[width=0.8\linewidth]{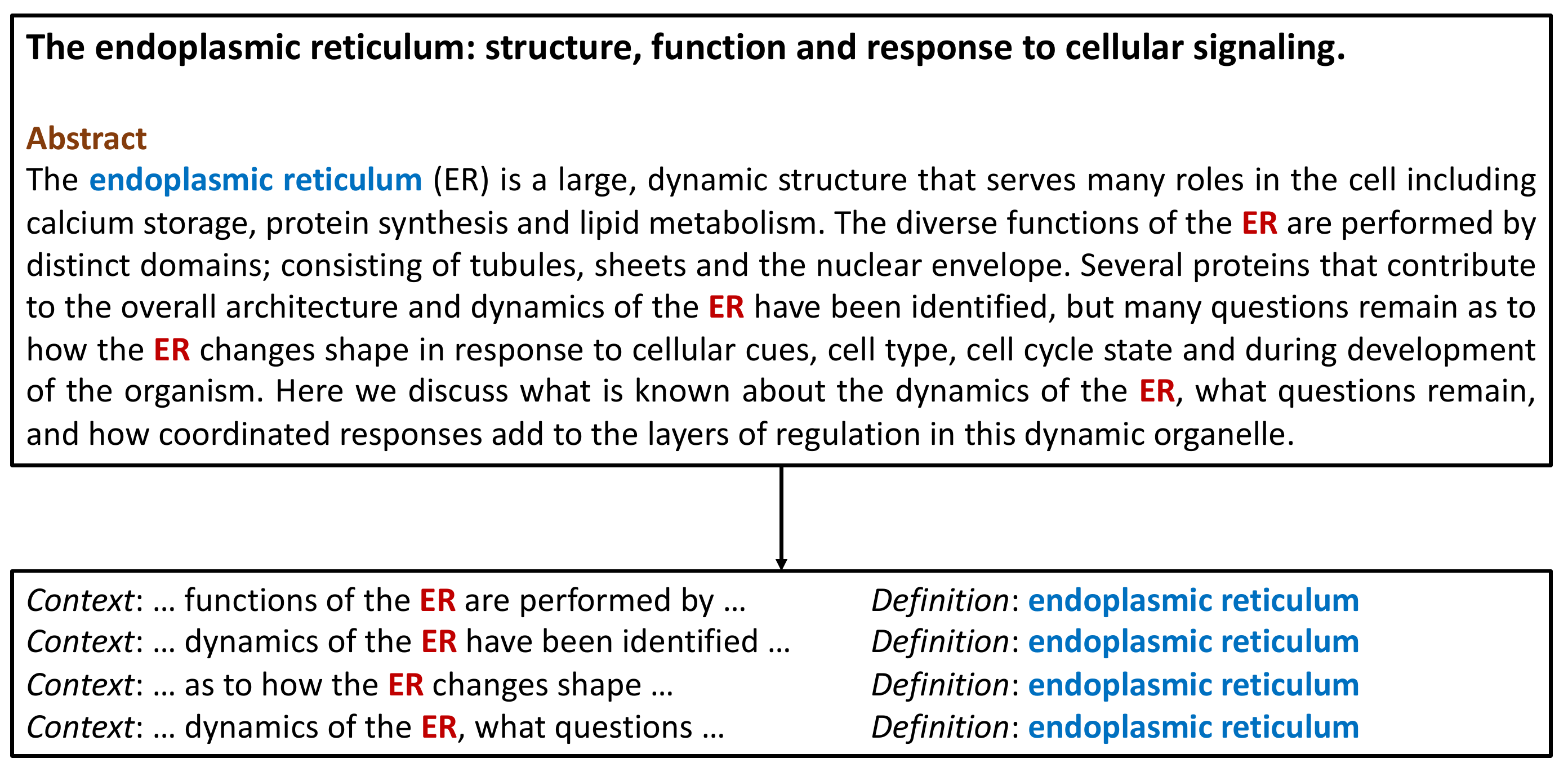}
    \caption{An example of automatically generated training instances for disambiguation from the abstract of \citet{schwarz2016endoplasmic}. In this case, we extract ``endoplasmic reticulum'' as the definition for all ER mentions in the abstract, and store those instances to the dataset.}
    \label{fig:dataset}
    \vspace{0.2cm}
\end{figure*}

Figure \ref{fig:dataset} shows our approach of automatically collecting disambiguation dataset. For each abstract, we first detect and extract the pattern of ``\textit{Definition} (\textit{Abbreviation})", e.g.: ``endoplasmic reticulum (ER)". Then we collect all the following sentences that contain the abbreviation, and label them with the definition.

This would generate a noisy label set due to the variations of writing the same definition (e.g.: emergency department and emergency departments). To group the same definitions together, we use MetaMap-derived MeSH terms \cite{demner2017metamap} as features of definitions and define the MeSH similarity between definition $a$ and definition $b$ as:
\[
s = \frac{\left|\mathcal{M}_a \cap \mathcal{M}_b\right|}{\sqrt{\left|\mathcal{M}_a\right|\left|\mathcal{M}_b\right|}}
\]
where $\mathcal{M}_a$ and $\mathcal{M}_b$ are the MeSH term sets of definition a and b, respectively. We group those definitions with high MeSH similarity and close edit distance by heuristic thresholds.

We collected 1970 abbreviations. However, due to the unsupervised nature of the collection process, some abbreviations are invalid or not ambiguous. For this, one biomedical expert\footnote{A post-doctoral fellow with a Ph.D. degree in biology.} filtered the abbreviations we found, based on 1) \textbf{Validity}: abbreviations should be biomedically meaningful; 2) \textbf{Ambiguity}: abbreviations should have multiple possible definitions, and prevalence of the dominant one should be $<99\%$. After the filtering, there are 950 valid ambiguous abbreviations. Their statistics are shown in Table \ref{tab:stat}. We split the instances of each abbreviation into training, development and test sets: If there is more than $10$k instances, we randomly select $1$k for both development and test sets. Otherwise, we randomly select 10\% of all instances for both development and test sets.
\begin{table*}[htbp]
\centering
\small
\begin{tabular}{lcccccc}
\toprule
\textbf{Statistic} & \textbf{Whole} & \textbf{Random} & \textbf{Imbalanced} & \textbf{Low-resources} & \textbf{Clinical} & \textbf{Human} \\
\midrule
\textbf{\# of all abbreviations} & 950 & 100 & 42 & 28 & 11 & 1\\
\textbf{Average \# of instances} & 8790.0 & 6564.3 & 19493.1 & 958.8 & 28642.8 & 8312.0\\
\textbf{Average \# of possible definitions} & 4.1 & 3.7 & 2.3 & 2.2 & 8.5 & 4.0\\
\textbf{Average \% of dominant definition} & 64.1 & 63.5 & 96.7 & 66.7 & 53.3 & 63.8\\
\bottomrule
\end{tabular}
\caption{Statistics of the automatically generated abbreviation disambiguation dataset and its subsets.}
\label{tab:stat}
\end{table*}

\subsection{BioELMo} \label{bioelmo}
BioELMo is a biomedical version of ELMo pre-trained on $10$ millions of PubMed abstracts \cite{jin2019probing}. It serves as a contextualized feature extractor in DECBAE: given an input sentence of $L$ tokens:
\[\text{input} = [t_1;t_2;...;t_L]\]
We use BioELMo to embed it to 
\[\mathbf{E} = [\mathbf{e_1};\mathbf{e_2};...;\mathbf{e_L}] \in \mathbb{R} ^ {L \times D}\]
where $\mathbf{e} \in \mathbb{R} ^ D$ is the token embedding and $D$ is the embedding dimension\footnote{Note that it's after scaling and averaging the 3 BioELMo layers using task-specific weights.}.

\subsection{Abbreviation-specific biLSTM Classifiers} \label{bilstm}
For each abbreviation, we train a specific biLSTM classifier, denoted as $\text{biLSTM}_i$ for abbreviation $i$. We feed the BioELMo representations of sentences containing abbreviation $i$ to $\text{biLSTM}_i$:
\[\text{biLSTM}_i(\mathbf{E}) = [\mathbf{h_1};\mathbf{h_2};...;\mathbf{h_L}] \in \mathbb{R} ^ {L \times 2H}\]
where $\mathbf{h} \in \mathbb{R} ^ {2H}$ is the concatenation of forward and backward hidden states of the biLSTM. We take as input the concatenated hidden states of the abbreviation $i$ (i.e. the ambiguous token) $\mathbf{h_a}$ and use several feed-forward neural network (FFN) layers with softmax output unit to predict its definition:
\[
p(\text{def}_k\,|\,\text{input}) \propto \text{exp}(\mathbf{w_{k}^T} \, \text{FFN}_{i}(\mathbf{h_a}))
\]
where $\mathbf{w_{k}}$ is the learnt weight vector corresponding to definition $k$, and $\text{def}_k$ is the $k$-th definition of abbreviation $i$ in our dataset. Similarly, we train FFN separately for different abbreviations.

\subsection{Training}
The weights of BioELMo are pre-trained and fixed, while the averaging weights and scaling factor of BioELMo embeddings are trained separately for each abbreviation along with the abbreviation-specific biLSTM classifiers. We use Adam \cite{kingma2014adam} to optimize the cross-entropy loss of the predicted label and ground-truth label.

\subsection{Inference} \label{infer}
At inference time, we denote the tokenized input sentence as $[t_1;t_2;...;t_L]$ and our ambiguous abbreviation set as $\mathcal{A}$. If $\exists t_j \in \mathcal{A}$, we run DECBAE to expand the $t_j$: First, we use BioELMo to compute the representations of all the input tokens to $\mathbf{E} = [\mathbf{e_1};\mathbf{e_2};...;\mathbf{e_L}]$. The trained biLSTM for abbreviation $t_j$, denoted as $\text{biLSTM}_{t_j}$, is retrieved and used to calculate the hidden states given the BioELMo embeddings of the input sentence:
\[\text{biLSTM}_{t_j}(\mathbf{E}) = [\mathbf{h_1};\mathbf{h_2};...;\mathbf{h_L}] \in \mathbb{R} ^ {L \times 2H}\]
Then $\mathbf{h_{t_j}}$, which is the concatenated hidden states of the ambiguous abbreviation $t_j$, is used for disambiguation through the trained abbreviation-specific FFN:
\[
\text{Definition}(t_j) = \text{def}_{\operatorname*{argmax}_{k} \mathbf{w_{k}^T} \, \text{FFN}_{t_j}(\mathbf{h_{t_j}})}
\]

\section{Experiments}
\subsection{Baseline Settings}
A trivial baseline is to predict the majority of definition for all cases, which could still lead to high accuracy in severely imbalanced datasets. We denote this method as \textbf{Majority}. We also test other baseline settings of different feature learning schemes. They are all followed by several FFN layers and a softmax output unit.

\textbf{Bag-of-words:} Following most of the previous works, we use bag-of-words features to represent the context by $\mathbf{c} \in \mathbb{R} ^ {\left|\mathcal{V}\right|}$, where $\left|\mathcal{V}\right|$ is the vocabulary size.

\textbf{BioELMo:} We take the BioELMo embeddings of the ambiguous abbreviations as input features.

\textbf{biLSTM:} We use biomedical w2v \cite{moen2013distributional} as word embeddings and train task-specific biLSTMs and use the hidden states of the ambiguous abbreviations as input features.

We also measure the \textbf{human performance}: due to limitation of resources, we just study single-expert performance on one sampled abbreviation. For this, the expert is shown with the test sentences, and asked to classify the ambiguous abbreviation to its possible definitions. An ensemble of experts will obviously generate better results, so our single-human results just represent the lower bound of human performance.

\subsection{Subset Settings}
We report the model performance on different subsets of our dataset. Statistics of those datasets are shown in Table \ref{tab:stat}.

\textbf{Random samples:} It's computationally expensive\footnote{ The rate-determining step is BioELMo due to its large size and recurrent nature.} and unnecessary to test the models on all 950 abbreviations. Instead, we use randomly sampled 100 abbreviations to represent the whole set.

\textbf{Imbalanced samples:} We define abbreviations whose dominant definitions have over 95\% frequency as imbalanced samples. Multi-label classification with imbalanced classes is considered as a hard machine learning task.

\textbf{Low-resources samples:} We define abbreviations that have less than 1k training instances as low-resources samples. It's motivated by the fact that most biomedical datasets are typically limited by scale, so models that can still perform well under low-resources settings have the potential to be applied in real world settings.

\textbf{Clinical samples:} Though our abbreviations are collected from PubMed abstracts, we have included 11 out of 13 of clinical ambiguous abbreviations mentioned in a previous work of clinical abbreviation disambiguation \cite{xu2012combining}. We also test our models on the subset of these 11 clinically related abbreviations. 

\textbf{Testing sample for human expert:} We test human performance on one abbreviation (DAT), due to limited resources. The statistics of DAT abbreviation expansion dataset are close to the averages of the whole dataset, as shown in Table \ref{tab:stat}. Possible definitions of DAT include: 1) Dopamine transporter (63.9\%); 2) Direct antiglobulin test (5.8\%); 3) Direct agglutination test (5.8\%); 4) Dementia of the Alzheimer type (24.5\%).

\subsection{Evaluation Metrics}
\begin{figure*}
    \centering
    \includegraphics[width=0.46\textwidth]{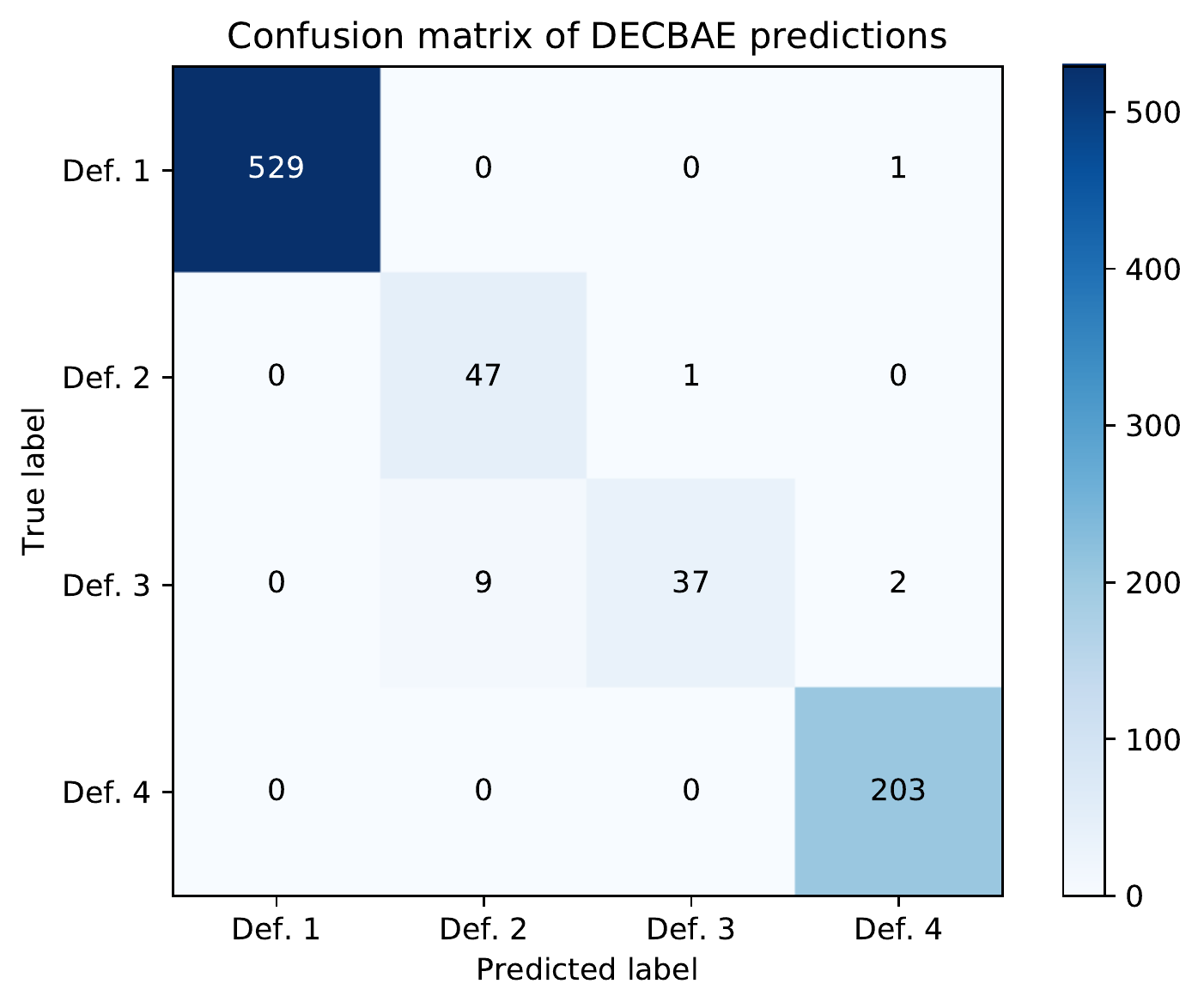} \hspace{0.15in}
    \includegraphics[width=0.46\textwidth]{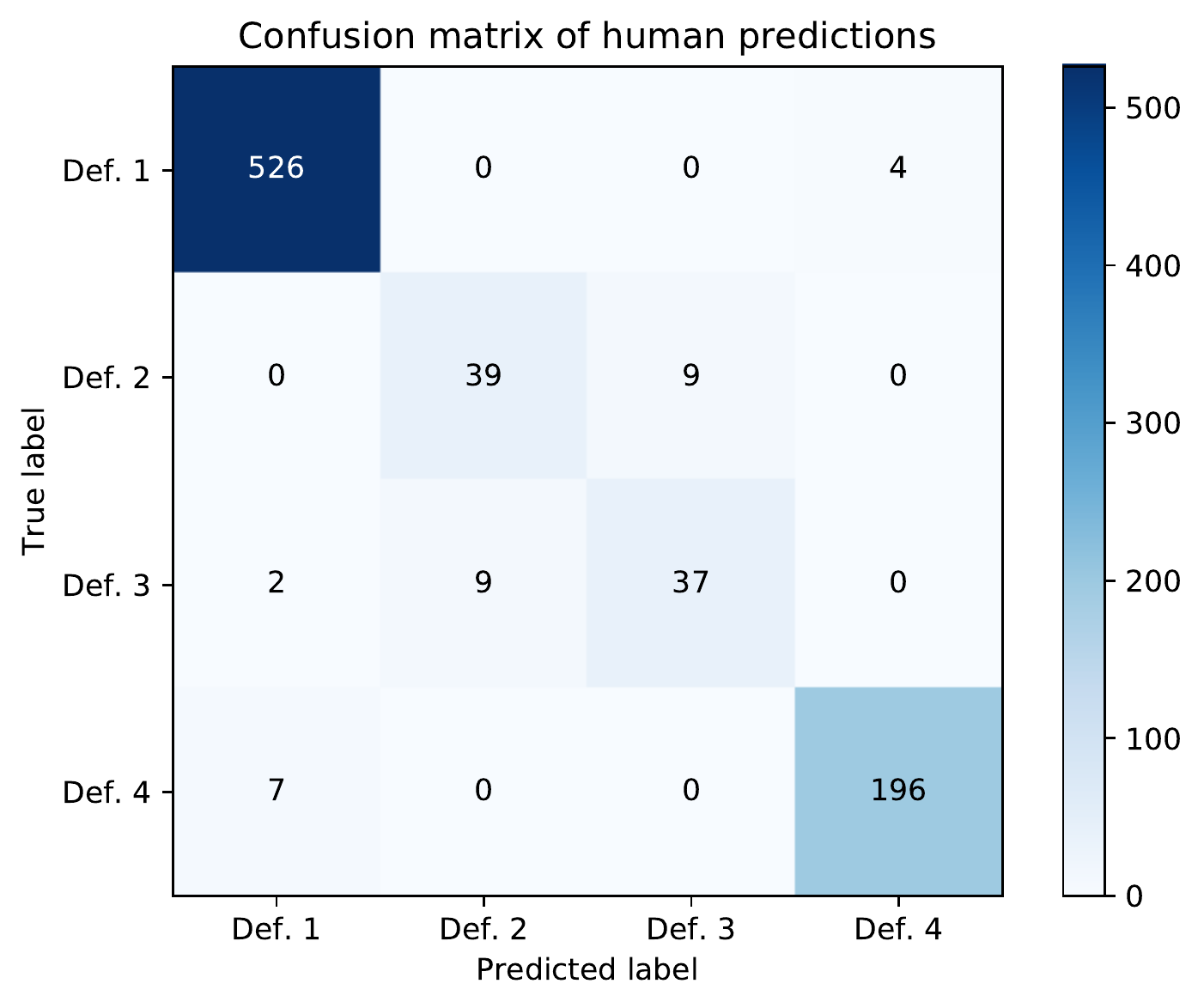}
    \caption{Confusion matrix for the predictions of DECBAE (left) and the human expert (right). Def. 1: dopamine transporter; Def. 2: direct antiglobulin test; Def. 3: direct agglutination test; Def. 4: dementia of the Alzheimer type.}
    \label{fig:cm}
\end{figure*}

\begin{table*}[htbp]
\centering
\small
\begin{tabular}{lcccccc}
\toprule
\textbf{Model} & \textbf{Random subset} & \textbf{Imbalanced subset} & \textbf{Low-resources subset} & \textbf{Clinical subset} & \textbf{Human testset} \\
\midrule
\textbf{Majority}\\
\midrule
\textbf{\hspace*{3mm}Accuracy} & 63.6\,$\pm$\,21.0$^\dagger$ & 96.7\,$\pm$\,1.0$^\dagger$ & 67.0\,$\pm$\,15.6$^\dagger$ & 53.3\,$\pm$\,25.7$^\dagger$ & 63.9\\
\textbf{\hspace*{3mm}Macro-F1} & 28.3\,$\pm$\,14.9$^\dagger$ & 45.4\,$\pm$\,8.8$^\dagger$ & 37.2\,$\pm$\,8.8$^\dagger$ & 12.0\,$\pm$\,10.6$^\dagger$ & 19.5\\
\textbf{\hspace*{3mm}Kappa Statistic} & 0.0\,$\pm$\,0.0$^\dagger$ & 0.0\,$\pm$\,0.0$^\dagger$ & 0.0\,$\pm$\,0.0$^\dagger$ & 0.0\,$\pm$\,0.0$^\dagger$ & 0.0\\
\midrule
\textbf{BoW-FFN}\\
\midrule
\textbf{\hspace*{3mm}Accuracy} & 84.4\,$\pm$\,11.2$^\dagger$ & 97.5\,$\pm$\,1.7$^\dagger$ & 89.6\,$\pm$\,7.5$^\dagger$ & 76.1\,$\pm$\,12.5$^\dagger$ & 84.3\\
\textbf{\hspace*{3mm}Macro-F1} & 73.1\,$\pm$\,17.1$^\dagger$ & 71.5\,$\pm$\,19.9$^\dagger$ & 83.4\,$\pm$\,14.6$^\dagger$ & 57.9\,$\pm$\,14.2$^\dagger$ & 71.9\\
\textbf{\hspace*{3mm}Kappa Statistic} & 63.8\,$\pm$\,25.3$^\dagger$ & 50.4\,$\pm$\,33.7$^\dagger$ & 71.1\,$\pm$\,24.8$^\dagger$ & 60.6\,$\pm$\,8.9$^\dagger$ & 69.6\\
\midrule
\textbf{BioELMo}\\
\midrule
\textbf{\hspace*{3mm}Accuracy} & 94.1\,$\pm$\,7.2$^\dagger$ & 96.3\,$\pm$\,15.3 & 98.1\,$\pm$\,2.7 & 91.1\,$\pm$\,8.4 & 97.1 \\
\textbf{\hspace*{3mm}Macro-F1} & 86.0\,$\pm$\,17.4$^\dagger$ & 81.3\,$\pm$\,23.5$^\dagger$ & 95.4\,$\pm$\,9.3 & 75.5\,$\pm$\,21.7 & 92.6 \\
\textbf{\hspace*{3mm}Kappa Statistic} & 86.1\,$\pm$\,19.8$^\dagger$ & 73.2\,$\pm$\,34.2$^\dagger$ & 93.2\,$\pm$\,10.8$^\dagger$ & 86.6\,$\pm$\,9.3 & 94.6 \\
\midrule
\textbf{biLSTM}\\
\midrule
\textbf{\hspace*{3mm}Accuracy} & 88.0\,$\pm$\,16.8$^\dagger$ & 98.0\,$\pm$\,1.9$^\dagger$ & 92.7\,$\pm$\,10.5$^\dagger$ & 88.2\,$\pm$\,8.2$^\dagger$ & 97.3 \\
\textbf{\hspace*{3mm}Macro-F1} & 77.1\,$\pm$\,26.0$^\dagger$ & 70.2\,$\pm$\,27.0$^\dagger$ & 82.9\,$\pm$\,24.5$^\dagger$ & 68.8\,$\pm$\,26.1 & 93.2 \\
\textbf{\hspace*{3mm}Kappa Statistic} & 69.3\,$\pm$\,37.2$^\dagger$ & 49.1\,$\pm$\,45.7$^\dagger$ & 70.4\,$\pm$\,41.5$^\dagger$ & 70.5\,$\pm$\,35.3 & 94.9 \\
\midrule
\textbf{DECBAE}\\
\midrule
\textbf{\hspace*{3mm}Accuracy} & \textbf{96.1\,$\pm$\,5.5} & \textbf{98.9\,$\pm$\,1.4} & \textbf{98.7\,$\pm$\,2.2} & \textbf{95.1\,$\pm$\,3.3} & \textbf{98.4} \\
\textbf{\hspace*{3mm}Macro-F1} & \textbf{91.7\,$\pm$\,13.2} & \textbf{87.2\,$\pm$\,17.8} & \textbf{98.3\,$\pm$\,3.5} & \textbf{83.0\,$\pm$\,21.9} & \textbf{93.9} \\
\textbf{\hspace*{3mm}Kappa Statistic} & \textbf{90.9\,$\pm$\,15.5} & \textbf{79.6\,$\pm$\,30.2} & \textbf{96.8\,$\pm$\,6.8} & \textbf{91.7\,$\pm$\,5.5} & \textbf{97.0} \\
\midrule
\textbf{Human Expert}\\
\midrule
\textbf{\hspace*{3mm}Accuracy} & -- & -- & -- & -- & 96.3\\
\textbf{\hspace*{3mm}Macro-F1} & -- & -- & -- & -- & 89.0\\
\textbf{\hspace*{3mm}Kappa Statistic} & -- & -- & -- & -- & 92.8\\
\bottomrule
\end{tabular}
\caption{Mean and standard deviation of model performance on different subsets. $^{\dagger}$Significantly lower than the corresponding metric of DECBAE. Significance is defined by $p<0.05$ in paired t-test. All numbers are in percentages. High deviations are expected due to the variety of abbreviations in each subset.}
\label{tab:results}
\end{table*}
We model abbreviation expansion as a multi-label classification task, and use the following metrics to measure the performance of different models:

\textbf{Accuracy:} Accuracy is defined as the proportion of right predictions in all predictions. Most of the definition labels are imbalanced, so accuracy could be misleadingly high for a trivial majority solution in these cases, thus may not reflect the real capability of models. 

\textbf{Macro-F1:} In multi-label classification, macro-F1 is calculated as an unweighted average of F1 score for each class. Class-wise F1 score is defined as follows:
\[\text{F1} = 2 \cdot \frac{\text{precision} \cdot \text{recall}}{\text{precision} + \text{recall}} \]
where precision and recall are calculated for each class.

\textbf{Kappa Statistic:} Cohen's kappa was originally introduced as a metric to measure inter-rater agreement \cite{cohen1960coefficient}. It can also be used to evaluate predictions of multi-label classification:
\[\kappa = \frac{p_o - p_e}{1 - p_e} \]
where $p_o$ is the observed agreement and in the case of classification $p_o=\text{accuracy}$, $p_e$ is the expected agreement which can be achieved by pure chance:
\[p_e = \sum_{c} p_c \hat{p_c}\]
$p_c$ and $\hat{p_c}$ refer to the proportion of class c in ground truth labels and predictions, respectively. Empirical results in Table \ref{tab:results} show that Kappa statistics are often lower than accuracy and macro-F1, and thus serving as a more distinctive metric for our task.

\subsection{Results}

In Table \ref{tab:results}, we report means and standard deviations of each model's performance on different subsets evaluated by the three metrics. In all subsets, DECBAE performs significantly better than most other models by large margins. A general trend of DECBAE $>$ BioELMo $>$ biLSTM $>$ BoW-FFN $>$ Majority conserves across subsets.

In the \textbf{Random} subset which represents the whole dataset, all metrics of DECBAE exceed $0.90$, setting very promising state-of-the-art performance despite the potential noise of the dataset.

In the \textbf{Imbalanced} subset where the most frequent definitions consist of over $95$\% of all the labels, a trivial Majority solution gets over $95$\% accuracy. However, for macro-F1 and kappa statistic, performance of the baselines drop dramatically while DECBAE can still generate decent results.

DECBAE and BioELMo alone remain robust in \textbf{Low-resources} setting. This is due to the transfer learning nature of BioELMo, which utilizes the knowledge encoded in the PubMed abstracts.

Our abbreviation expansion dataset covers roughly 85\% of clinical abbreviations mentioned in \citet{xu2012combining}. On this \textbf{Clinical} subset, DECBAE gets pretty good results and vastly outperform other baselines despite its variety in possible definitions (8.5 possible definitions per abbreviation, as shown in Table \ref{tab:stat}).

On the testset for human performance (i.e.: abbreviation expansion for DAT), DECBAE and even some neural baselines outperform single human expert.

\begin{table*}[htbp]
\centering
\small
\begin{tabular}{cccc}
\toprule
\textbf{Test sentence} & \textbf{Label} & \textbf{Human} & \textbf{DECBAE} \\
\midrule
\makecell{The reduction of the number of different segments in \textbf{DAT} compared to controls and\\ patients suffering from depression may be helpful for differential diagnosis.} & Def. 4 & Def. 1 & \textbf{Def. 4} \\
\makecell{Reliance on objective brain phenotype measures, for example, those afforded by brain \\imaging, might critically improve detection of \textbf{DAT} genotype-phenotype association.} & Def. 1 & \textbf{Def. 1} & Def. 4 \\
\makecell{\textbf{DAT} was more commonly positive among BO incompatible (21.5\% in BO vs. 14.8\% \\in AO , P=0.001) and black (18.8\% in blacks vs. 10.8\% in nonblacks , P=0.003) infants.} & Def. 2 & Def. 3 & \textbf{Def. 2} \\
\makecell{NPY-LI showed a significant reduction in \textbf{DAT} but not in FTD.} & Def. 4 & Def. 1 & \textbf{Def. 4} \\
\makecell{The study included 122 healthy subjects, aged 18-83 years, recruited in the multicentre \\`ENC-\textbf{DAT}' study (promoted by the European Association of Nuclear Medicine).} & Def. 1 & Def. 4 & \textbf{Def. 1} \\
\bottomrule
\end{tabular}
\caption{Some samples of errors made by the human expert and DECBAE. Def. 1: dopamine transporter; Def. 2: direct antiglobulin test; Def. 3: direct agglutination test; Def. 4: dementia of the Alzheimer type.}
\label{tab:err}
\end{table*}

\section{Analysis}
In Fig. \ref{fig:cm}, we use confusion matrices to visualize the differences between DECBAE or the human expert and the ground truth labels, for disambiguation of abbreviation ``DAT''. The high agreement level between human expert predictions and the automatically assigned labels indicates that our pipeline of collecting the abbreviation disambiguation dataset is valid.

In general, both DECBAE and the human expert perform well in the task, with only few misclassifications. Specifically, DECBAE, and even other neural baselines like biLSTM and BioELMo, outperform the human expert in all metrics. Compared to DECBAE, the human expert is more likely to misclassify direct agglutination test with direct antiglobulin test (9 v.s. 1), and misclassify dementia of the Alzheimer type with dopamine transporter (7 v.s. 0). We show several instances of human and DECBAE's errors in Table \ref{tab:err}. 

One limitation of this work is that we just test DECBAE on our automatically collected dataset. Since the proposed model can also be used on other biomedical abbreviation expansion datasets as well, evaluating on other datasets like MSH WSD is a clear future work to do.

Another potential direction for improvement is to accelerate the inference speed. Currently DECBAE uses BioELMo for embedding and abbreviation-specific biLSTM for classification, resulting in two recurrent models in total. Our results show that just BioELMo with several FFN layers also generates decent results, so in some cases we might use only BioELMo as a compromise for faster inference.

\section{Conclusion}
We present DECBAE, a state-of-the-art biomedical abbreviation expansion model on the automatically collected dataset from PubMed. The results show that, with only minimum expert involvement, we can still perform well in such a domain-specific task by automatically collecting training data from a large corpus and utilize embeddings from pre-trained biomedical language models. 

\section{Acknowledgement}
We are grateful for the annonymous reviewers of BioNLP 2019 who gave us very insightful comments and suggestions. J.L. is supported by NLM training grant 5T15LM007059-32.

\bibliography{acl2019}
\bibliographystyle{acl_natbib}
\end{document}